%% file: ERarxiv.tex
\documentclass{article} % For LaTeX2e
\usepackage{arxiv,times}

\input{math_commands.tex}

\usepackage[utf8]{inputenc} % allow utf-8 input
\usepackage[T1]{fontenc}    % use 8-bit T1 fonts
\usepackage{hyperref}       % hyperlinks
\usepackage{url}            % simple URL typesetting
\usepackage{booktabs}       % professional-quality tables
\usepackage{amsfonts}       % blackboard math symbols
\usepackage{nicefrac}       % compact symbols for 1/2, etc.
\usepackage{microtype}      % microtypography
\usepackage{xcolor}         % colors

\usepackage{amsmath}
\usepackage{graphicx}
\usepackage{multirow}
\usepackage{subcaption}
\usepackage{amssymb}
\usepackage{enumerate,enumitem}
\usepackage{wrapfig}
\usepackage{algorithm, algpseudocode}

\usepackage{setspace}
\usepackage{caption}
\usepackage{colortbl}

\def\mb{\mathbf}
\def\mbb{\mathbb}
\def\mc{\mathcal}

\title{Energy Ring: Stabilizing Feature Energy \\ for Reliable Test-Time Discovery}

\author{
  Tianle Liu\textsuperscript{1}, Fan Lyu\thanks{Corresponding authors.}~~\textsuperscript{2}, Tingliang Feng\textsuperscript{3}, Chenggong Ni\textsuperscript{4}, Tianqi Wang\textsuperscript{1}, Zhang Zhang\textsuperscript{5},\\ ~\textbf{Fuyuan Hu}\textsuperscript{$*$ \textbf{1}}, \textbf{Rui Yao}\textsuperscript{\textbf{6}}, \textbf{Tao Zhou}\textsuperscript{\textbf{7}}\\
  \textsuperscript{1}School of Electronics and Information Engineering, Suzhou University of Science and Technology\\
  \textsuperscript{2}Computer Vision Center, Universitat Autònoma de Barcelona\\
  \textsuperscript{3}College of Computer Science, Northwest University \\
  \textsuperscript{4}School of Intelligence Science and Technology, Nanjing University \\
  \textsuperscript{5}New Laboratory of Pattern Recognition, Institute of Automation, Chinese Academy of Sciences \\
  \textsuperscript{6}School of Computer Science and Technology, China Unviersity of Mining and Technology\\
  \textsuperscript{7}School of Computer Science Engineering, North Minzu University\\
  \texttt{tianleliu@post.usts.edu.cn}\\
  \texttt{fanlyu@cvc.uab.cat} ~~~~~~
  \texttt{fuyuanhu@mail.usts.edu.cn}
}

\iclrfinalcopy % Uncomment for camera-ready version, but NOT for submission.
\begin{document}

\maketitle

\begin{abstract}
Test-Time Discovery (TTD) aims to recognize known classes and detect novel ones while the model is already deployed, operating directly on a continual unlabeled test-time stream. Existing TTD methods are overly conservative, making known and unknown features collapse together and difficult to separate during test-time adaptation.
In this paper, we revisit TTD from an energy perspective and find that the natural energy gap between known and unknown classes is unstable and easily collapses. We therefore propose Energy Ring to stabilize this gap by anchoring the two types of classes on separate energy rings.
A lightweight adapter enlarges the norm-based gap while preserving known-class directions, and a perturbation consistency module forms compact clusters for unknown classes. During testing, energy-guided confidence weighting improves prediction, and a stability-based rectification mechanism corrects unreliable updates. Experimental results show that Energy Ring significantly improves TTD performance against existing TTD methods.
Our code will is available at \url{https://github.com/LeTianL-TT/ER-for-TTD}.
\end{abstract}

\section{Introduction}
\label{sec:intro}

Novel Class Discovery (NCD)~\citep{pu2023dynamic, wu2023metagcd} aims to identify new categories from unlabeled data using knowledge from known classes. However, it separates the testing and discovery phases, assuming a fixed unlabeled set available in advance. Such a static setting is unsupported in open environments, where new classes appear continuously after deployment. 
Recently, Test-Time Discovery (TTD)~\citep{lyu2025test} was proposed to detect and learn novel classes while the model is already deployed. 
Unlike traditional NCD, TTD operates in a continual, unlabeled stream where both known and unknown samples arrive sequentially under test-time constraints, while discovering and testing in real time. This capability is essential for real-world applications where new concepts may emerge unexpectedly after model deployment~\citep{tan2026logic, mitigateocl, fewshotcil}.
Existing TTD methods usually take a conservative strategy to avoid false discovery of new classes and to protect performance on known ones.
As shown in Fig.~\ref{fig:recent}, HM~\citep{lyu2025test} freezes the backbone so that pretrained features stay fixed. This prevents drift on known classes but also blocks useful updates for new classes. DAA~\citep{liudaa} adds a learnable adapter that allows updates, yet forces all feature vectors to have the same norm. This improves stability but removes a key source of variation, which may lead to crowding and confusion across unknown and known classes.
In this paper, we revisit the TTD problem from an energy view. Prior studies~\citep{objectsphere, gu2025activation,park2023norm, ding2026vd} suggest that feature energy, measured by the norm of a feature, reflects class-specific information. In many cases, known classes have higher energy and unknown classes have lower energy. This natural gap can be used as a cue to separate old and new classes at test time, reducing interference and improving discovery.
\textit{However, the energy of unknown classes is easy to turn chaotic, and the boundary between high and low energy erodes, yielding the unreliability of the whole system.} 

Motivated by this, we aim to \textit{amplify} and \textit{stabilize} the energy difference between known and unknown classes. %增加一句描述，说明这个ER的由来，我们希望TTD过程中xxxx，像是ER一样。
Features are expected to be distributed at varying heights during TTD just like concentric rings, with each category forming distinct clusters, thereby facilitating easier discrimination and discovery.
In this paper, we therefore propose \textbf{Energy Ring}, a TTD framework that can anchor known classes on a high-energy ring and unknown classes on a low-energy ring. 
Specifically, we introduce a lightweight energy ring adapter that enlarges separation while preserving feature directions of known classes to retain prior knowledge. In addition, we enforce consistent perturbations to encourage compact and stable clustering of unknown samples along the energy ring. During testing, the learned energy-ring structure enables query classification via energy-weighted confidence, while a high-perturbation priority rectification mechanism identifies unstable predictions and corrects them, further improving the accuracy and stability of the energy rings.
Our method improves agreement performance by 3.33\% compared to DAA~\citep{liudaa}.
In summary, our contributions are: 

\begin{figure}[t]
    \centering    
    \includegraphics[width=1\linewidth]{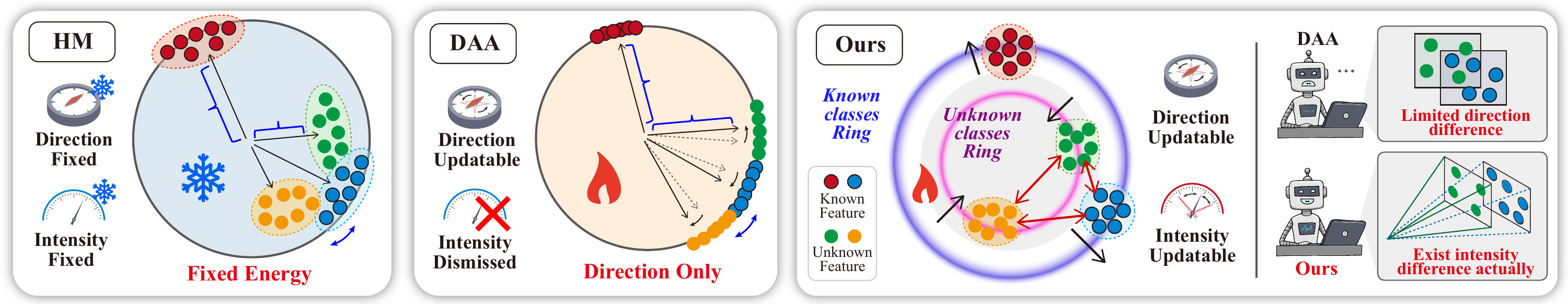}
    \vspace{-15px}
    \caption{
    Energy comparisons among existing TTD methods and our method. HM keeps the backbone frozen and classifies based on fixed spatial energy. DAA collapses the original energy intensity onto a single hypersphere. Energy Ring compresses known and unknown classes onto different energy rings, yielding spatial-energy difference and retaining separability. 
    }
    \label{fig:recent}
    \vspace{-15px}
\end{figure}

\begin{enumerate}[label=(\arabic*),left=0pt,itemsep=0pt,parsep=0pt]
\vspace{-5px}
\item We propose Energy Ring, a TTD framework that stabilizes the inherently fragile energy gap by anchoring known and unknown classes on separate energy rings.
\item We design a lightweight energy ring adapter and a perturbation-consistency module that jointly enlarge the energy gap, preserve known-class directions, and produce compact, stable clusters for unknown classes.
\item Extensive experiments demonstrate the effectiveness of Energy Ring, showing clear improvements over prior TTD methods.
\end{enumerate}

\section{Related Work}
\label{sec:related}

\textbf{Novel Category Discovery (NCD)}~\citep{han2020automatically, zhang2023promptcal, zhong2021neighborhood} tackles the problem of recognizing classes that never appear in the training labels. Standard NCD pipelines first learn a transferable representation from the labelled data, then cluster the unlabelled stream to reveal new groups. DTC~\citep{han2019learning} exploits semantic anchors from related classes to reduce clustering ambiguity; GCD~\citep{vaze2022generalized} and PromptCCD~\citep{cendra2025promptccd} replace the traditional k-means stage with contrastive learning prompting for purer prototypes. On-the-fly Category Discovery (OCD)~\citep{du2023onfly, zheng2024PHE} pushes NCD toward real-time operation by hashing incoming images and assigning them to the nearest Hamming bucket. OCD, however, is designed only for discovery whose backbone remains frozen, no fine-tuning on the new classes is performed. NCD methods treat discovery and adaptation as two separate stages.
% which will break down when the model must respond in real-time.

\noindent
\textbf{Test-Time Training (TTT)}~\citep{reftta, sun2020test, tan2026towards, lyu2025mitigating} is a paradigm that adapts a pre-trained model to distribution shifts while the model is already deployed. Instead of relying on a frozen model, TTT injects self-supervised objectives and uses the test stream itself to update the backbone on the fly. Representative methods such as TTT++~\citep{liu2021ttt++} and TTT-MAE~\citep{tttmae} show that this on-line alignment significantly improves accuracy under covariate shift~\citep{refreplay1, myyicme, Tan2026iclr}. Yet the vast majority of TTT work assumes the model is only asked to preserve performance on the known classes, leaving it blind to categories that emerge after deployment.

\noindent
\textbf{Test-Time Discovery (TTD)} unifies the two desiderata: retain accuracy on old classes while instantly detecting and learning new ones during inference. The first TTD implementation, pioneered by Lyu et al.~\citep{lyu2025test}, keeps the network frozen and instead maintains an external memory built with Locality-Sensitive Hashing. By avoiding any gradient update, the method sidesteps catastrophic forgetting, but it also forfeits the possibility of refining the feature space when old and new classes overlap. DAA retains the adaptation spirit of TTT while embracing the discovery mandate of NCD: it updates only a small set of adapters in real time, thereby sharpening the decision boundary between known and emerging classes without erasing previous knowledge.

\section{Methods}
\label{sec:method}

\subsection{TTD Definition and Energy-based Discovery}
\begin{figure}[t]
    \centering    
    \includegraphics[width=0.9\linewidth]{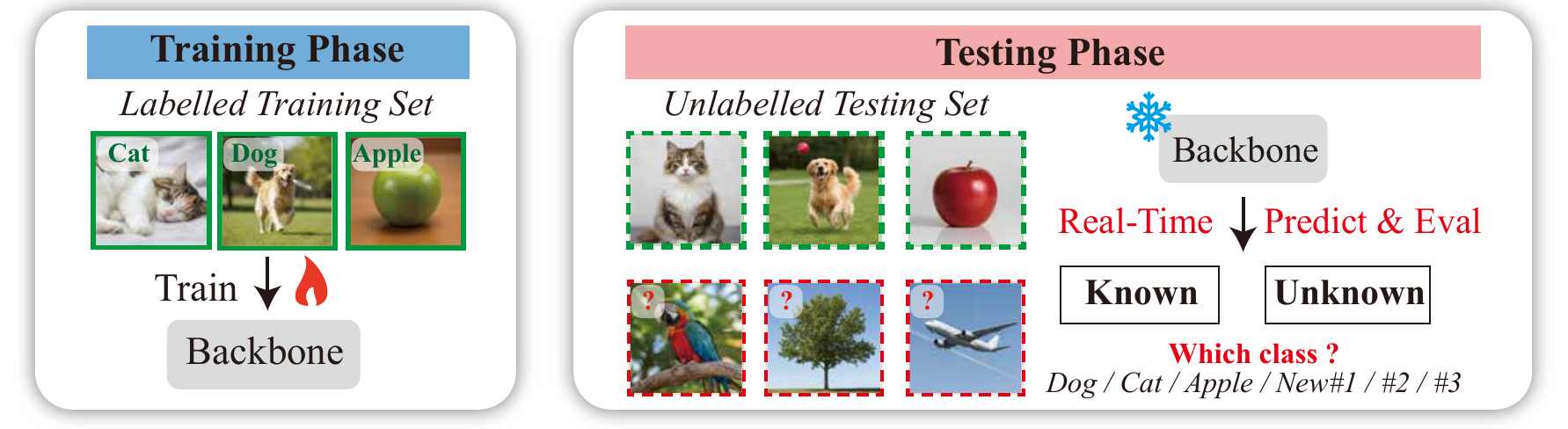}
    \vspace{-8px}
    \caption{
    Task setting of TTD. In TTD, a model is pretrained on data containing only known classes. During deployment, the test data may include both known and unknown classes, requiring the model to predict known classes and discover novel ones.
    }
    \label{fig:defin}
    \vspace{-17px}
\end{figure}

% TTD needs to immediately determine which class each sample belongs to (including newly discovered unknown classes, rather than relying on post-clustering classification), and the results need to be evaluated in real-time during testing.
As shown in Fig.~\ref{fig:defin}, TTD aims to recognize known classes and discover unknown ones in a continuous unlabeled stream after model deployment.
Formally, let \(\mathcal{D}_{\text{train}} = \{(x_i, y_i)\}_{i=1}^{N}\) denote the training set, where \(x_i \in \mathcal{X}\) is input image and \(y_i \in \mathcal{Y}_{\text{kn}}\) is its label from the known class set \(\mathcal{Y}_{{kn}}\). 
During testing, the model encounters a test stream \(\mathcal{D}_{\text{test}} = \mathcal{D}_{\text{kn}} \cup \mathcal{D}_{\text{un}}\), where \(\mathcal{D}_{\text{un}}\) may contain unknown classes that were never seen during training. TTD requires model to satisfy two objectives simultaneously. 
First, \textit{novel class discovery and learning}, where inputs \( x \in \mathcal{D}_{\text{test}}^{\text{un}} \) must be identified as belonging to unseen classes and assigned new labels to distinguish them from known-class samples.
Second, \textit{unified classification}, where all inputs \( x \in \mathcal{D}_{\text{test}} \) are accurately classified into either a known class \( y \in \mathcal{Y}_{\text{kn}} \) or a novel class \( y \in \mathcal{Y}_{\text{seen}} \), with \( \mathcal{Y}_{\text{seen}} \) denoting the set of dynamically discovered novel classes.

% Unlike Novel Category Discovery (NCD) which assumes static labeled/unlabeled splits, TTD operates in a continual, label-free stream where both known and unknown samples arrive sequentially.

\begin{wrapfigure}[11]{r}{0.6\textwidth}
    \centering
    \vspace{-15px}
    \includegraphics[width=\linewidth]{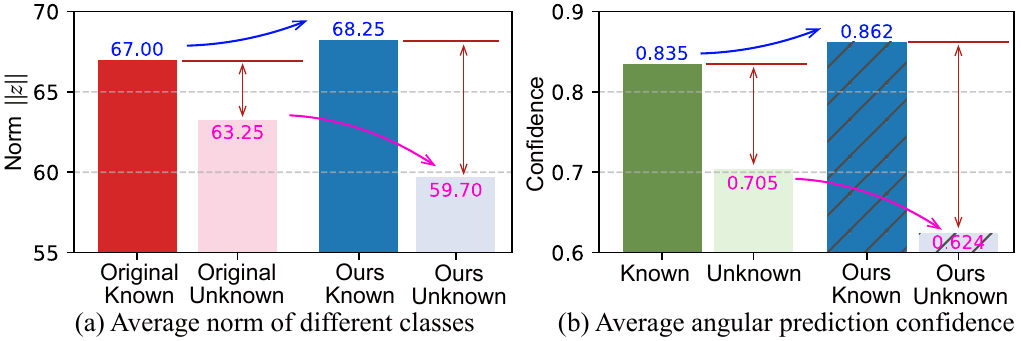}
    \vspace{-18px}
    \caption{Comparison of norm and angular prediction for test samples. Our method further expands the differences of energy in both intensity and direction.
    }
    \label{fig:normmethod}
    % \vspace{-10pt}
\end{wrapfigure}
% \subsection{Preliminary: Energy-based Discovery}
In this paper, we study TTD from an energy view. During testing, the model receives a mixture of known and unknown samples and must make decisions based on the learned representation \(z=F(x)\). 
For known classes, the supervised objective explicitly optimizes the dependency between visual features and their corresponding class identifiers. 
In contrast, unknown classes are absent from training and their visual information is not explicitly organized or aligned with the learned class-discriminative space.
This distinction often manifests in the magnitude of learned representations. 
Discriminative training tends to amplify feature responses along directions that are useful for separating known classes, producing more confident and higher-energy representations for in-distribution samples. 
% Unknown samples, lacking direct supervision and alignment to the learned class anchors, are more likely to activate these discriminative directions weakly, leading to lower feature energy or smaller feature norms on average. 
This motivation is consistent with prior OOD detection and novelty discovery studies showing that known samples often exhibit stronger activation energy, while unknown samples tend to produce lower feature responses~\citep{gu2025activation,objectsphere,park2023norm, ni2025maintaining}.
Thus, we use the feature norm \(\|z\|_2\) as a soft indicator of representation confidence, rather than as a strict separation criterion between known and unknown classes. 
This preliminary creates a natural difference in energy, reflected in both the norm and direction of features. As illustrated in Fig.~\ref{fig:normmethod}, known classes usually appear with higher energy and unknown ones with lower energy.
% We also report the average feature norms before Energy Ring adaptation with different backbone in the Appendix.
Simulated features have lower average norms and therefore provide an appropriate low-energy initialization cue.
However, this separation is unstable as their purpose is to expose ERA to ambiguous boundary regions rather than to imitate the semantics of future categories. The norms of different classes overlap, the angular differences are weak, and as more novel classes arrive, these patterns may drift and cause confusion between old and new categories.
These observations motivate us to design a method that enlarges and stabilizes the energy difference so that known and unknown classes remain distinguishable during test-time discovery.
% More validation can be seen in Appendix.

% \begin{figure}[h]
%     \centering    
%     \includegraphics[width=0.9\linewidth]{figs/normmethod.pdf}
%     \vspace{-10px}
%     \caption{
%     Comparison of norm and angular for test samples. Our method further expands the differences of energy in both intensity and direction.
%     }
%     \label{fig:normmethod}
%     \vspace{-15px}
% \end{figure}

% During testing, the model receives both known and unknown samples, and must rely on the feature representation $z=F(x)$ to make decisions. Since known classes have been learned during training, their features carry class information, while unknown classes satisfy Mutual Information $I(Z;Y_{\text{un}})=0$. Feature extractor yielding finite second‑order moment with isotropic covariance, it tends to the maximum entropy distribution:
% \begin{equation}
% \lim_{I\to 0}p(z|x)=\arg\max_{p}H(p)\propto\exp\left(-{\|z\|_2^2}/{(2\sigma^2)}\right),
% \end{equation}
% resulting in the L2-norm of the feature representation $\|Z_{\text{new}}\|_2$ being lower than $\|Z_{\text{old}}\|_2$, that is, the new class norm is smaller than the known class norm, and the difference monotonically increases with the entropy difference. Many prior works on OOD detection and novelty discovery also report that discriminative training amplifies feature energy for known classes while unknown samples exhibit lower energy \citep{gu2025activation,objectsphere,park2023norm}.

We therefore explicitly construct two non-overlapping energy rings, keeping known classes at higher energy ring and unknown classes at lower energy ring. We introduce a lightweight \textbf{Energy Ring Adapter (ERA)} and (i) construct energy rings through warm up after training the backbone model to achieve the effect of expanding the energy difference among known and unknown classes while maintaining the energy direction of the learned old classes; (ii) enhance the stability of the unknown classes energy ring through perturbations and prepare for rectification. During testing, we (iii) utilize the constructed energy rings for energy weighted prediction to improve the accuracy of discovery and classification and (iv) conduct test-time training to correct and optimize the energy rings based on the predicted results and high-perturbation priority rectification mechanism.

\begin{figure*}[t]
    \centering    
    \includegraphics[width=\linewidth]{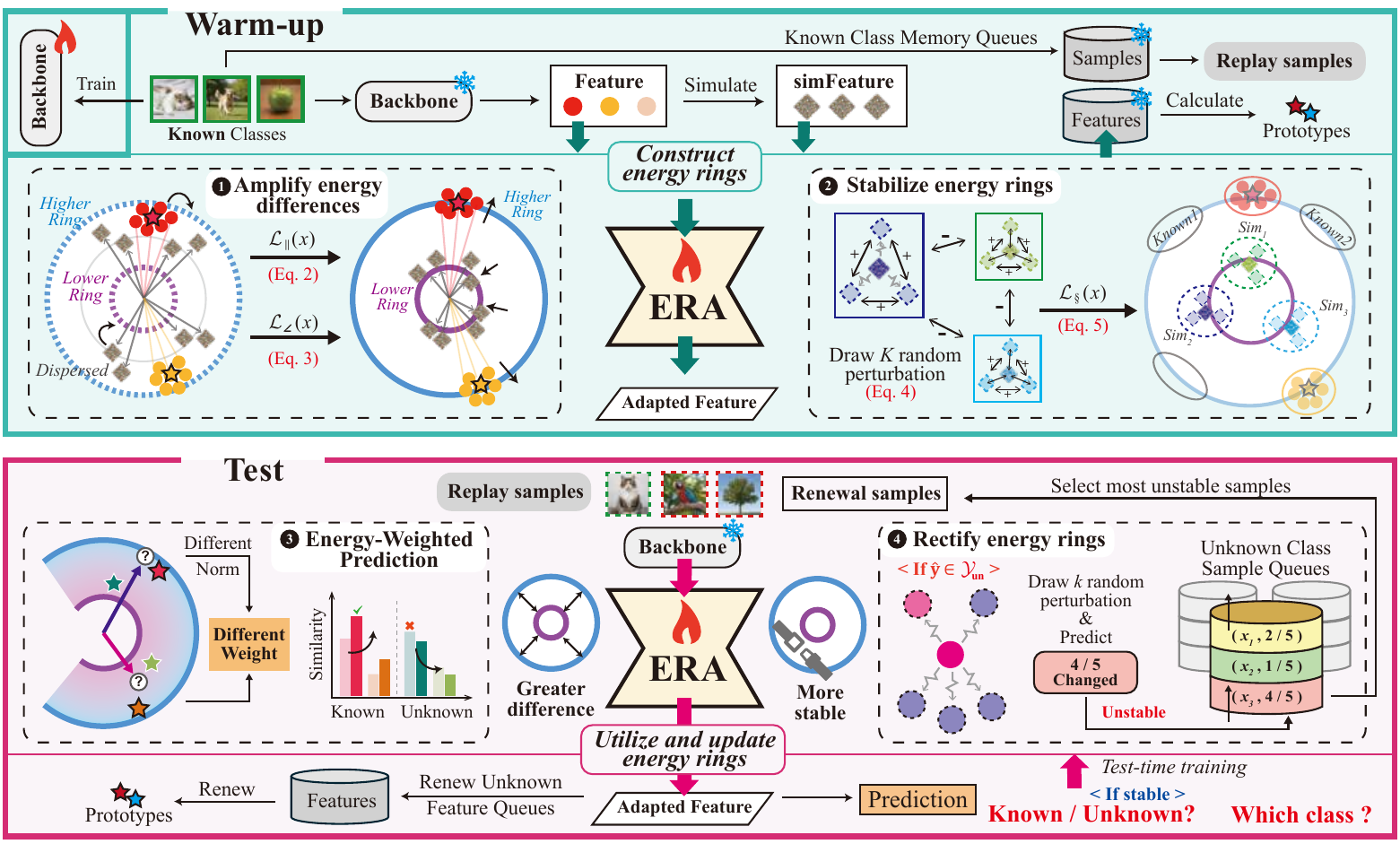}
    \vspace{-10px}
    \caption{
     Method Schema. ERA projects features onto concentric energy rings: high-energy ring for known classes and low-energy ring for unknowns, together with perturbation-consistency during training. During testing, we conduct energy-weighted prediction, while selectively rectifying high-perturbation samples to refresh memory queues and update ERA to rectify energy rings. 
    }
    \label{fig:setting}
    \vspace{-10px}
\end{figure*}

\subsection{Preconstruct Energy Rings through Warm-up}

% To let the model recognise unknown classes without ever clustering them, we pre-train the lightweight Adapter with a three-loss curriculum that relies on simulated unknowns and keeps every constraint in absolute scale (no L2 normalisation).  
% The whole stage updates only the Adapter parameters; the backbone stays frozen.

Following DAA~\citep{liudaa}, we also adopt a warm-up strategy to enlarge the gap between known and unknown classes before testing. 
Unlike DAA, which collapses all features onto a single hypersphere and removes energy differences, our goal is to preserve and further widen the natural gap observed earlier. Since known and unknown classes already show different energy scales, warm-up encourages them to lie on different energy levels, making later discovery more reliable. To achieve this, we introduce an adapter ERA inserted into the pre-trained backbone, which transforms the original feature $\mathbf{r}(x)=f(x)$ into an adapted representation $\mathbf{z}=\text{ERA}(\mathbf{r}(x))$. Specifically, we simulate unknown samples during warm-up by mixing pairs of features with a coefficient $\lambda$ drawn from a Beta distribution and adding Gaussian noise, forming a synthetic feature $\tilde{\mathbf{r}}(x)=\lambda\mathbf{r}(x)+(1-\lambda)\mathbf{r}(x')+\mathbf{n}$. These mixed features serve as proxies for unknown classes~\citep{liudaa,zhang2017mixup} and help stabilize and enlarge the energy separation before test-time discovery.
Below, we construct the energy rings during warm-up through \textbf{three aspects}: \textit{1) constraining known and unknown classes onto different energy tracks (norm constraint), 2) maintaining separability for known classes (direction constraint), and 3) enhancing the stability of the energy rings under perturbation}.

%不要写加噪，参考文献，就像某篇

% DAA only has limited effect because it collapses the original radial depth into a single hypersphere. Therefore, we need to warm-up based on spatial energy, so that features can be layered spatially, making it easier to classify during testing phase.
% Given an input image $x$, the backbone produces a feature vector $\mathbf{r}(x)=f(x) \in \mathbb{R}^{D}$, which is then transformed by \red{ERA} into an adapted representation \(\mathbf{z} = A(\mathbf{r}(x))\). We also followed DAA to simulate Unknown-Class samples to increase training diversity by combining Gaussian noise and the Mixup technique~\citep{zhang2017mixup}. 
% Given two difference samples $x$ and $x'$, the synthetic feature $\mathbf{\tilde{r}}(x)=\lambda\,\mathbf{r}(x)+(1-\lambda)\,\mathbf{r}(x')+\mathbf{n}$, where $\mathbf{n}$ denotes a vector of Gaussian noise with a mean of zero and a certain covariance matrix. The mixing coefficient $\lambda$ is drawn from a Beta distribution.

% Previous TTD method DAA fails because known and unknown features collapse into the same spherical space. We employ ERA to forces them into well-separated energy rings and the distinction among unknown classes is made more stable by perturbing the simulated samples. 
First, we enforce a norm constraint that pushes known samples onto an outer ring and unknown samples onto an inner ring, creating a clear spatial gap between them:
% The core design of DAA lies in keeping the backbone that contains known classes unchanged, and entrusting the discovery and learning of new classes to the constructed DAA module.  
% To achieve this, we first propose a discrepancy-amplifying warm-up strategy for DAA update before the testing phase. 
%写成期望形式，排一行，把x和r写进自变量，总分形式写有三个loss，first要干啥，then要干啥，
% \noindent \textbf{Radial Expansion in Spatial Energy-Ring}. 
% \begin{equation}
% \mathcal{L}_{||}(x)=\frac{1}{|\mathcal{B}_{\text{kn}}|}\sum_{i}\bigl(\|A(\mathbf{r}(x_i))\|_2-r_\text{kn}\bigr)^2
% +
% \frac{1}{|\mathcal{B}_{\text{un}}|}\sum_{j}\bigl(\|A(\tilde{\mathbf{r}}(x_j))\|_2-r_\text{un}\bigr)^2,
% \end{equation}
\begin{equation}
\mathcal{L}_{||}(x)=
\Bigl(\|\text{ERA}(\mathbf{r}(x))\|_2-(\mb{1}_{\text{kn}}(x)\,r_{\text{kn}}+\mb{1}_{\text{un}}(x)\,r_{\text{un}})\Bigr)^2,
\label{eq1}
\end{equation}
where \(\mb{1}(x)\) is an indicator function, if $x$ is a known class, \(\mb{1}_{\text{kn}}=1,\mb{1}_{\text{un}}=0\) and vice versa. \(r_\text{kn}\) and \(r_\text{un}\) denote known and unknown classes energy ring. 

Second, to preserve previously learned knowledge, ERA must not alter the semantic direction of known-class features. We maintain feature direction by maximizing cosine similarity between the original and adapted features:
\begin{equation}
\mathcal{L}_{\angle}(x)=1-\cos\bigl(\mathbf{r}(x),\,\text{ERA}(\mathbf{r}(x))\bigr).
\label{eq2}
\end{equation}

Third, even with spatial separation and direction preservation, the simulated unknown samples generated by mixup and noise remain random and unstable. To enhance the consistency of unknown classes, we introduce a perturbation-based constraint. For each adapted unknown feature \(\tilde{\mathbf{z}}=\text{ERA}(\tilde{\mathbf{r}}(x))\), we create \(K\) norm-preserving perturbations by sampling random directions \(\boldsymbol{\delta}\sim\mathcal{N}(0,\mathbf{I})\), projecting them onto the tangent space of \(\tilde{\mathbf{z}}\), and scaling them to an angular radius \(\varepsilon\). The perturbed features therefore remain on the same energy ring:
\begin{equation}
\mathbf{z}_b = \mathbf{\tilde{z}} + \varepsilon\,
\frac{\boldsymbol{\delta} - \bigl(\boldsymbol{\delta}^{\!\top}\mathbf{\tilde{z}}\bigr)\mathbf{\tilde{z}}/\|\mathbf{\tilde{z}}\|^{2}}
{\bigl\|\boldsymbol{\delta} - \bigl(\boldsymbol{\delta}^{\!\top}\mathbf{\tilde{z}}\bigr)\mathbf{\tilde{z}}/\|\mathbf{\tilde{z}}\|^{2}\bigr\|}.
\end{equation}
% For each sample \( i \) in a batch $b$, we build a soft positive set \(\mathcal{P}_b = \bigl\{ \mathbf{z}_b^{(i)} \mid i=1,...,K \bigr\}\) which are different perturbations of same feature and a soft negative set \(\mathcal{N}_b = \bigl\{ \mathbf{z}_{b'}^{(i)} \mid b'\neq b,\;i=1,...,K \bigr\}\) which are different features, which can pull together small, norm-preserving perturbations of the same simulated unknown feature, and pushing apart perturbations that originate from different simulated feature, thereby sharpening the inter-class dispersion without destroying the original norm information:
We treat the \(K\) perturbations of the same feature as soft positives \(\mathcal{P}_b = \{ \mathbf{z}_b^{(i)} \mid i=1,...,K \}\) and perturbations from different features as soft negatives \(\mathcal{N}_b = \{ \mathbf{z}_{b'}^{(i)} \mid b'\neq b,\;i=1,...,K \}\). This encourages perturbations of the same unknown sample to stay close, while pushing those from different samples apart, improving the stability and separability of unknown classes without altering norm information:
\begin{equation}
\mathcal{L}_{\S}=
\frac{1}{K}\sum_{b}\max\!\Bigl(0,\,
\|\mathbf{z}_{b}^{(0)}\!-\!\bar{\mathcal{P}}_{b}\|
-\|\mathbf{z}_{b}^{(0)}\!-\!\bar{\mathcal{N}}_{b}\|\Bigr).
\label{eq4}
\end{equation}

The final warm-up loss combines all three terms:
\begin{equation}
\mathcal{L}_{\text{warmup}}=\mathcal{L}_{||}+\mathcal{L}_{\angle}+\mathcal{L}_{\S}.
\end{equation}

% Because the feature space is not L2-normalised, the loss operates on the absolute geometry inherited from the energy-ring pre-training, directly optimising the separation of perturbation clouds while leaving the norm information intact.
% \textbf{Total Pre-Training Objective.}  

%三个小标题去掉，要写出这里和能量环关系不要叫train loss

\subsection{Energy Ring Guided Prediction and Discovery}
\label{subsec:joint}
Warm-up establishes the initial energy rings for known and unknown classes. We now move to the testing phase and describe how these energy rings guide prediction and discovery. To provide energy information, similar to DAA, we maintain memory queues for both known and undiscovered classes.
% During warm-up, the energy of features is constrained to widen the spatial differences of features. In the testing phase, we use the spatial distribution characteristics to assist in class prediction and classification. We also followed DAA to maintain the memory queues. However, we prioritized correcting unstable samples based on the stability of features, which can make the framework's knowledge more accurate and the memory queue better able to describe categories correctly.
%要和能量环挂钩，怎么用能量环，图也要画变化，怎么构建怎么利用怎么更新
%都不要用小标题
% \noindent\textbf{Energy-Weighted Confidence.}
% During testing phase, the model receives an unlabeled mini-batch  
% \(\mathcal{B}_{\text{test}}=\{x_{i}\}_{i=1}^{B}\).
For a test image \(x\), its ERA representation \(\mathbf{z}=\text{ERA}(\mathbf{r}(x))\). 
% Due to the known class being more inclined towards higher energy, if the norm of the feature is closer to the higher energy loop $r_k$, then the sample is more likely to belong to the known class. On the contrary, it tends to belong to the unknown class. 
We first compute an energy coefficient \(\rho\) based on its position between the two energy rings:
\begin{equation}
\rho=\frac{\|\mathbf{z}\|_{2}-r_\text{un}}{r_\text{kn}-r_\text{un}}\in[0,1].
\end{equation}
% where $\|\mathbf{z}\|_{2}$ denotes the norm of feature $\mathbf{z}$. 
Let \(\boldsymbol{\mu}_{c}\) be the prototype of class \(c\). Features closer to the outer ring should receive slightly higher similarity toward known classes. Given a small amplitude \(\alpha\in(0,1)\), we adjust cosine similarities within a narrow range \([1-\alpha,\,1+\alpha]\), so the energy prior nudges but does not dominate the prediction:
\begin{equation}
\tilde{s}_{c}=
\begin{cases}
\cos(\mathbf{z},\boldsymbol{\mu}_c)\cdot [1+\alpha(2\rho-1)],& c\in\mathcal{Y}_{\text{kn}}\\
\cos(\mathbf{z},\boldsymbol{\mu}_c)\cdot [1-\alpha(2\rho-1)],& c\in\mathcal{Y}_{\text{seen}}.
\end{cases}
\end{equation}
%sc写cos，w放进下面,两个字母不能连在一起
% Therefore, known classes distributed on the outer ring will be better classified as known classes due to their higher confidence weighting. The confidence of unknown classes in the inner loop will decrease compared to known classes, making them easier to discover and classified as unknown classes. Based on the max similarity scores, we apply a confidence threshold \(\gamma\) to determine whether a test sample $x$ belongs to a seen class or an unseen class.
This weighting increases confidence for known classes on the outer ring and decreases confidence for unknown classes on the inner ring. We then use a confidence threshold \(\gamma\) to decide whether a test sample belongs to an existing class or a new one:
\begin{equation}
\hat{y} = \begin{cases} 
\text{arg max}_{c\in(\mc{Y}_{\text{kn}}\cup\mc{Y}_{\text{seen}})}(\tilde{s}_{c}), & \text{if } \max_c \tilde{s}_{c} > \gamma, \\
\text{new discovered class}, & \text{otherwise}.
\end{cases}
\end{equation}

% \noindent\textbf{Perturbation Consistency Check.}
% For each sample judged as an unknown class, we need to check its stability and repeatedly use the perturbation mechanism from the warm-up phase. Draw \(k\) random directions \(\boldsymbol{\delta}_{k}\sim\mathcal{N}(0,\mathbf{I})\) and project them at \(\mathbf{z}_{\hat{y}}\in\mathcal{Y}_{\text{un}}\). Let \(\mathbf{z}^{(k)}=\mathbf{z}+\boldsymbol{\delta}_{k}\) and recompute \(\tilde{s}_{c}^{(k)}\) and make predictions again for each. Then count the number of unchanged categories and calculate the stability coefficient $S$:
To ensure the reliability of an unknown prediction, we adopt a stability recheck strategy. 
For a sample predicted as an unknown class, we reuse the perturbation mechanism used in warm-up. 
We draw \(k\) random directions \(\boldsymbol{\delta}_{k}\sim\mathcal{N}(0,\mathbf{I})\), project them at the feature \(\mathbf{z}_{\hat{y}}\), and form perturbed features \(\mathbf{z}^{(k)}=\mathbf{z}+\boldsymbol{\delta}_{k}\). 
For each perturbed feature we recompute the similarity scores \(\tilde{s}_{c}^{(k)}\) and obtain a new prediction. 
The stability of the original prediction is then measured by
\begin{equation}
S=\frac{1}{k}\sum_{i=1}^{k}\mb{1}\left[\arg\max_{c}\tilde{s}_{c}^{(i)}=\hat{y}\right]
\end{equation}
where $\mb{1}(\cdot)$ is the indicator function. For each unknown-class sample, we store the tuple \((x_{c}^{i}, S^{i})\) in the queue. Samples with the lowest stability are selected for rectification and added to the test-time training, helping reduce prototype drift and improving the accuracy of class representations. And the model will treat this sample as the first instance of this new seen class. The image feature \(\mathbf{z}\) will be used as the initial prototype for this class, denoted as \(\boldsymbol{\mu}_{\hat{y}}\). Subsequent samples will be compared to all known and seen prototypes, including this new \(\boldsymbol{\mu}_{\hat{y}}\).

\subsection{Test-Time Training to Stabilize Energy Rings}  
\label{sec:test_time_train}

During testing phase, ERA not only produces energy-aware features but also continues to learn from the streaming batch. The process requires real-time updating of the adapter, strengthening the dual ring spatial structure, while preventing the forgetting of old knowledge, and performing memory correction to selectively update the most unstable unknown samples to refine the class margin. For known classes, we continue to use lower energy ring loss and direction-preservation loss, which penalizes deviations between the original and adapted features and constrains known class features on the larger energy ring to ensure that the ERA does not alter the feature representations significantly. 
Contrary to known classes, stable unknown classes that have been evaluated are processed using larger energy ring loss and contrastive loss, which encourages ERA to better adapt to new knowledge of unknown classes.
Specifically, given a test batch data $\mc{B}$, the test-time training with ERA can be represented by two kinds of loss functions.
The first kind of loss is the self-supervised learning loss via the pseudo labels:
% \begin{equation}
% \begin{array}{l}
% \mathcal{L}_{\text{kn}}(\mathcal{B},q)=
% \mbb{E}_{\substack{x\in\mathcal{B}\\\hat{y}=q\in\mathcal{Y}_{\text{kn}}}}
% \Bigl[\,
% \mathcal{L}_{\text{energy}}(x\mid\hat{y})
% +
% \mathcal{L}_{\text{dir}}(x\mid\hat{y})
% \Bigr],
% \\
% \mathcal{L}_{\text{un}}(\mc{B},q) =  \mbb{E}_{\substack{x\in\mathcal{B}\\\hat{y}=q\in\mathcal{Y}_{\text{un}}\\S>\kappa}}~\Bigl[\mathcal{L}_{\text{energy}}(x|\hat{y}, S)+\mathcal{L}_{\text{cont}}(x|\hat{y},S)\Bigr],
% \end{array}
% \end{equation}
\begin{equation}
\begin{array}{l}
\mathcal{L}_{\text{kn}}(x)=
\mbb{E}_{\substack{\hat{y}\in\mathcal{Y}_{\text{kn}}}}
\Bigl[\,
\mathcal{L}_{||}(x\mid\hat{y})
+
\mathcal{L}_{\angle}(x\mid\hat{y})
\Bigr],
\;
\mathcal{L}_{\text{un}}(x) =  \mbb{E}_{\substack{\hat{y}\in\mathcal{Y}_{\text{un}}\\S>\kappa}}~\Bigl[\mathcal{L}_{||}(x|\hat{y}, S)+\mathcal{L}_{\text{cont}}(x|\hat{y},S)\Bigr],
\end{array}
\end{equation}
where $\mathcal{L}_{\text{cont}}$ represents Contrastive Loss~\citep{mao2023cross}. The second kind represents the replay for known and the renewal for seen classes via retrain samples in memory. Specifically, for every discovered class $p$ we retrieve the samples with the lowest $S_{p}^{i}$ (highest instability) regularly, and these samples would be reprocessed, re-evaluated, and update ERA again:
\begin{equation}
\begin{array}{c}
\mathcal{L}_\text{kn}^{\text{replay}}(x) = \mbb{E}~\mathcal{L}_{||}(x)+\mathcal{L}_{\angle}(x),
\quad
\mathcal{L}_\text{un}^{\text{renewal}}(x) =
\mbb{E}~\mathcal{L}_{||}(x)+\mathcal{L}_{\text{cont}}(x).
\end{array}
\end{equation}

The overall testing phase loss is shown below:
% \begin{equation}
% \begin{array}{c}
% \mathcal{L}_\text{test}(\mc{B}) = \lambda_1\cdot\mbb{E}_{q\in\mc{Y}_\text{kn}}
% \left(
% \mathcal{L}_{\text{kn}}(\mc{B},q) + \mathcal{L}_\text{kn}^{\text{replay}}(q)\right) 
% \\
% + \lambda_2\cdot\mbb{E}_{q\in\mc{Y}_\text{seen}}
% \left(
% \mathcal{L}_{\text{un}}(\mc{B},q)
% +\mathcal{L}_\text{un}^{\text{renewal}}(q)\right),
% \end{array}
% \end{equation}
\begin{equation}
\begin{array}{c}
\mathcal{L}_\text{test}(x) = \lambda_1\cdot\mbb{E}_{\hat{y}\in\mc{Y}_\text{kn}}
\left(
\mathcal{L}_{\text{kn}}(x|\hat{y}) + \mathcal{L}_\text{kn}^{\text{replay}}(x)\right) 
+ \lambda_2\cdot\mbb{E}_{\hat{y}\in\mc{Y}_\text{seen}}
\left(
\mathcal{L}_{\text{un}}(x|\hat{y})
+\mathcal{L}_\text{un}^{\text{renewal}}(x|\hat{y})\right),
\end{array}
\end{equation}
where \(\lambda_1\) and \(\lambda_{\text{2}}\) are hyperparameters that balance the contributions of each loss term.

\section{Experiment}
\label{sec:exp}

%后面四个去掉，单独放一个小的

%阈值法不要了，caption里面缩到两行，caption去掉数据集

\subsection{Experimental Details}
\noindent\textbf{Datasets}.
We conduct our experiments based on four benchmark datasets.
CIFAR100 (C100)~\citep{cifar100krizhevsky2009learning} is a commonly used classic dataset. 
Tiny ImageNet~\citep{tinyle2015tiny} is a subset of ImageNet~\citep{deng2009imagenet}.
Caltech-UCSD Birds-200-2011 (CUB)~\citep{cubwah2011caltech} and FGVC-Aircraft~\citep{aircraft2013fine} are complex fine-grained datasets.
All these datasets are split into known and unknown classes (7:3).
The model is trained on the known training set and tested on the mixture of known and unknown test sets.
% Statistics of the benchmark datasets and data splits are shown in Table~\ref{tab:dataset}.
Because the dataset is used for discovery, we follow the names of four transformed datasets from HM~\citep{lyu2025test}: CIFAR100D, CUB-200D, Tiny-ImageNetD, and AircraftD.
% More dataset details can be seen in Appendix.

% \vspace{-20px}
\noindent\textbf{Implementation details}.
In our implementation, we build our method on the prompt-based method L2P~\citep{wang2022learning}, which employs a ViT-B/16 backbone~\citep{khan2022transformers} following the pretraining procedure of NCD and GMP work. We employed the contrastive loss of the GCD literature when we fine-tune the pretrained model on the known classes, using SGD optimizer and cosine decay learning rate scheduler with an initial learning rate of 0.1 and minimum learning rate of 0.0001, and weight decay of 0.00005. All input images are resized to 224 × 224 and augmented to match the pretrained backbone settings. %We build our framework with PyTorch on a single NVIDIA RTX 4090 GPU.
\begin{table*}[t]
\centering
\setlength{\tabcolsep}{4pt}
\caption{Comparisons on four datasets. Real-time evaluation reflects the accumulated performance across all test batches, while post evaluation reassesses all test samples after testing. \textbf{Bold} and \underline{underlining} denote the best and second-best results.}
\vspace{-7px}
\resizebox{\textwidth}{!}{
\begin{tabular}{cl|rrr|rrrr}
\toprule
\multirow{2}{*}{} &  \multirow{2}{*}{\textbf{Method}} &  \multicolumn{3}{c|}{\textbf{Real-time Evaluation}} &
  \multicolumn{4}{c}{\textbf{Post Evaluation}} \\
% \cmidrule{3-10}
  &           & \multicolumn{1}{c}{KA$\uparrow$} & \multicolumn{1}{c}{TA$\uparrow$} &  \multicolumn{1}{c|}{CA$\uparrow$} &  \multicolumn{1}{c}{KA$\uparrow$} & \multicolumn{1}{c}{TA$\uparrow$} &  \multicolumn{1}{c}{CA$\uparrow$} &  \multicolumn{1}{c}{KF$\downarrow$} \\
\midrule
\multirow{7}{*}{\rotatebox[origin=c]{90}{\textbf{CIFAR100D}}}                                        
&  L2P~\citep{wang2022learning} &  59.93±2.15 &  8.57±2.49 &   43.10±4.30 &   50.53±7.25 &  9.60±1.50 &  27.39±2.11 &   27.85±7.21 \\
& DP~\citep{wang2022dualprompt}        &66.09±1.01       & 8.80±1.69 & \underline{48.34}±6.78 &  56.19±2.00 & 8.68±2.06 & 28.93±2.25 & 29.06±1.99    \\
& GMP~\citep{cendra2025promptccd}       & 72.77±1.20     & 7.37±0.88 & 42.26±4.54 &  67.21±2.53 & 13.04±2.59 &  27.10±1.77 & 17.69±2.53  \\
& PHE~\citep{zheng2024PHE}     & 72.35±1.12 & 16.74±1.05 &  37.34±1.31 & 70.12±1.07 & 14.55±\textbf{1.01} & 24.65±1.35  & \textbf{2.12}±0.92  \\
& HM~\citep{lyu2025test}      & 79.17±\textbf{0.13} &  21.13±\textbf{0.62} &  \textbf{56.37}±1.42 &   \textbf{80.73}±1.59 & 31.03±1.24 &  34.81±\underline{1.22}  &  \underline{3.41}±1.49 \\
& DAA~\citep{liudaa}      &  \underline{80.81}±0.44 & \underline{31.94}±0.98  & 45.38±\textbf{0.88}  & 80.27±\textbf{0.60} & \underline{35.63}±\underline{1.11} &  \underline{37.31}±\textbf{1.20}
& 4.10±\underline{0.60}  \\
\rowcolor{orange!20} & Ours    & \textbf{81.79}±\underline{0.42} & \textbf{33.31}±\underline{0.77} & 47.42±\underline{1.04} & \underline{80.40}±\underline{0.66} & \textbf{37.23}±1.18 & \textbf{38.25}±1.34 & 4.00±\textbf{0.50}  \\
\midrule
\multirow{7}{*}{\rotatebox[origin=c]{90}{\textbf{CUB200D}}}        

& L2P~\citep{wang2022learning}       & 46.22±1.53 &  9.01±\underline{0.87} &   55.37±7.79 &    31.97±3.35 &  4.75±\underline{0.73} &   24.48±1.65 &  42.29±3.14  \\
& DP~\citep{wang2022dualprompt}         & 53.69±1.24 &  43.68±2.20 &   45.68±5.88 &  63.37±3.27 &  45.94±0.91 &  7.92±1.10&  5.85±3.11  \\
& GMP~\citep{cendra2025promptccd}       &  62.97±1.33 &  46.44±1.87 & 47.99±4.34 &  58.11±3.00 &  48.02±1.20&  10.31±1.45&  5.46±2.77  \\
& PHE~\citep{zheng2024PHE}  & 57.29±1.23 & 32.68±1.66 &  57.22±2.75 & 54.77±1.75 & 27.92±\textbf{0.43} & 21.36±1.36 & \textbf{2.18}±0.71  \\
& HM~\citep{lyu2025test}      & 66.20±\underline{0.55} &  \textbf{58.30}±2.37 &   43.33±6.10 &  64.42±\underline{0.65} &  \textbf{65.28}±1.78&  37.25±4.90 &    4.07±\underline{0.47}  \\
& DAA~\citep{liudaa}      & \textbf{68.09}±\textbf{0.33} & 54.49±1.03 & \textbf{58.34}±\textbf{1.35}  & \underline{66.26}±\textbf{0.39} & 64.13±2.23 &  \underline{44.65}±\textbf{0.88} & 3.60±\textbf{0.39}   \\
\rowcolor{orange!20} &  Ours   & \underline{67.10}±1.48 & \underline{57.63}±\textbf{0.47} & \underline{58.19}±\underline{1.42} & \textbf{67.38}±1.82 & \underline{64.23}±1.78 & \textbf{45.53}±\underline{1.05} & \underline{3.00}±0.52  \\
\midrule
\multirow{7}{*}{\rotatebox[origin=c]{90}{\textbf{T-ImageNetD}}}
& L2P~\citep{wang2022learning}       & 46.25±1.41 &  7.79±2.92 &    53.55±6.47 &  29.50±3.77 &  10.38±2.42 &  23.28±\textbf{0.79} &  47.97±3.77  \\
& DP~\citep{wang2022dualprompt}         &46.51±\textbf{0.58} &  6.41±0.93 &  58.27±6.10 & 28.53±3.33 &  9.63±2.10 &  26.80±1.38 &  47.57±3.32   \\
& GMP~\citep{cendra2025promptccd}       &62.47±1.40 &  6.25±1.72 &  58.02±4.29 &  63.95±2.04 &  15.64±2.63 &   26.31±2.33 &    16.86±2.04   \\
& PHE~\citep{zheng2024PHE}  & 63.28±1.29 & 13.54±1.20 & 70.33±3.80 &  58.39±\textbf{1.14} & 12.80±\textbf{1.05} & 19.42±1.45 &  1.88±\textbf{0.35}  \\
& HM~\citep{lyu2025test}      & 75.31±1.31 &  16.04±\underline{0.76} &  \textbf{73.81}±2.67 &   74.94±2.20 &  16.23±\underline{1.24} &  \underline{37.43}±1.30 &  \textbf{1.15}±2.18  \\
& DAA~\citep{liudaa}      & \underline{76.38}±\underline{0.82} & \underline{24.10}±0.80 &  71.70±\underline{2.02}  & \textbf{75.50}±\underline{1.96} & \underline{23.17}±1.35  & 36.41±\underline{1.12} & 2.39±1.56   \\
\rowcolor{orange!20} &  Ours      &  \textbf{76.72}±1.25 & \textbf{25.23}±\textbf{0.71} & \underline{72.60}±\textbf{1.43} &\underline{75.28}±1.98 & \textbf{24.33}±1.38 & \textbf{37.93}±1.49 & \underline{1.50}±\underline{0.95}  \\

\midrule
\multirow{7}{*}{\rotatebox[origin=c]{90}{\textbf{AircraftD}}} 
& L2P~\citep{wang2022learning}       & 43.21±1.08 & 20.11±1.20  & 35.52±1.93 & 40.17±2.34 & 21.00±1.21 & 27.21±\textbf{1.33} & 15.04±2.10  \\
& DP~\citep{wang2022dualprompt}         &40.18±1.10 & 21.88±1.23 &  34.80±\underline{1.79} &42.53±\textbf{1.46} & 20.22±\textbf{0.96} & 28.15±1.74 & 12.68±1.74  \\
& GMP~\citep{cendra2025promptccd}       &41.26±\underline{1.04} & 21.12±\underline{0.67} &  35.10±2.50 & 32.56±1.75 & 21.25±\underline{1.02} & 26.34±\underline{1.48} & 22.65±2.25  \\
& PHE~\citep{zheng2024PHE}  & 45.82±\textbf{1.02} & 22.56±1.05  & \underline{41.72}±2.13 & 47.28±2.88 & 25.09±1.76 & 24.38±1.67 & 7.93±1.13  \\
& HM~\citep{lyu2025test}      & \underline{50.35}±1.08 & \underline{28.01}±0.85 & 40.23±2.22  & \textbf{49.77}±1.94 & \textbf{29.80}±1.25 & 28.74±1.53 & \underline{5.44}±1.56  \\
& DAA~\citep{liudaa}      &  49.37±1.46 & 27.79±\textbf{0.31} & 40.51±\textbf{1.05} & 48.87±1.73 & 28.37±1.69 & \underline{28.99}±1.69 & 5.61±\underline{1.04}   \\
\rowcolor{orange!20} &  Ours      &  \textbf{50.88}±1.42 & \textbf{28.49}±0.76 & \textbf{41.90}±1.91 &\underline{49.42}±\underline{1.66} & \underline{29.79}±1.50 & \textbf{30.79}±1.55 & \textbf{5.20}±\textbf{0.96}  \\
\bottomrule
\end{tabular}
}
\label{tab:major}
\vspace{-10px}
\end{table*}

\noindent\textbf{Evaluation metrics}.
% Evaluating the performance of models in the context of Test-Time Discovery (TTD) requires a comprehensive set of metrics that can accurately assess both known and novel class classification capabilities. Our evaluation framework is structured to provide insights into the model's real-time performance as well as its overall accuracy after the test phase is completed.
Following HM, we use the following metrics.
For unknown classes, we use two agreement metrics: (1) \textit{True-label Agreement ratio} (\textbf{TA}). This metric measures the maximum proportion of samples from a given true class that are predicted as the same class; and (2) \textit{Cluster Agreement ratio} (\textbf{CA}). This metric measures the maximum proportion of samples from a given predicted cluster that have the same true label.
% \begin{equation}
%     \text{TA} = \mbb{E}_{c \in \mc{Y}^\text{GT}_\text{seen}} \frac{1}{|\mc{D}^\text{test}_c|}\max_{p\in\mc{Y}_\text{seen}}\left(\sum\nolimits_{x\in\mc{D}^\text{test}_c}\mb{1}[\hat{y}(x)=p]\right),
% \end{equation}
% \vspace{-5px}
% \begin{equation}
%     \text{CA} = \mbb{E}_{p \in \mc{Y}_\text{seen}} \frac{1}{|\mc{C}^\text{test}_p|}\max_{c\in\mc{Y}^\text{GT}_\text{seen}}\left(\sum\nolimits_{(x,y)\in\mc{C}^\text{test}_p}\mb{1}(y=c)\right).
% \end{equation}
% where $\mb{1}(\cdot)$ is the indicator function (1 if true, 0 otherwise). 
We also provide some clustering metrics that traditional NCD methods use, including Hungarian Cluster Accuracy (HCA)~\citep{meilua2003comparing}, Adjusted Rand Index (ARI)~\citep{rand1971objective}, Normalized Mutual Information (NMI)~\citep{mcdaid2011normalized} and V-Measure~\citep{rosenberg2007v}.
% (1) \textbf{HCA} measures the clustering accuracy by optimal one-to-one mapping between predicted clusters and true labels using the Hungarian algorithm. 
% (2) \textbf{ARI} quantifies the similarity between the predicted clustering assignments and the true labels while adjusting for chance.
% (3) \textbf{NMI} assesses the mutual dependence between predicted and true labels by shared information between the two distributions.
% (4) \textbf{VM} simultaneously constrains the purity and coverage of the clusters through the harmonic mean.
For known classes, we also employ two key metrics to comprehensively assess the model’s performance: \textit{Known Accuracy} (\textbf{KA}) and \textit{Known Forgetting} (\textbf{KF}). KA measures the traditional classification accuracy of the model on known classes while KF quantifies the degree of performance degradation on known classes over time.
% For more details of the metrics, see \app{Appendix}.

\subsection{Major Comparisons}
In this paper, we first compare our methods with naive thresholding-based training-free methods. When exceeding the threshold, the naive method will be considered to have discovered a new class. And we compare with some training-required methods including L2P~\citep{wang2022learning}, DP~\citep{wang2022dualprompt}, and GMP~\citep{cendra2025promptccd} as these methods update prompts like TTA and CL methods. We also compare with the recent PHE~\citep{zheng2024PHE} method for OCD and HM~\citep{lyu2025test}, DAA~\citep{liudaa} for TTD.

\noindent\textbf{Comparisons on TA \& CA}.
The analysis of our experimental results, as illustrated in Table~\ref{tab:major}, offers key insights into the performance of various methods in the context of Test-Time Discovery (TTD). Training-based methods, which update model parameters upon encountering new classes, tend to suffer from performance degradation. This decline is largely due to their immediate adaptation to novel classes, resulting in lower TA and CA, as well as an increased risk of catastrophic forgetting. On the other hand, training-free methods, such as HM, avoid parameter updates but has restricted capacity to refine its representations in response to novel data. While DAA incorporates a lightweight module that can be updated which only focuses on energy direction and the amplification of inter-class differences remains limited. Our proposed method strikes a more balanced effectiveness among these challenges, achieving more consistent performance across all four datasets. These results demonstrate that our approach enhances the network’s adaptability in discovering novel classes without compromising the retention of previously acquired knowledge.

\noindent \textbf{Comparison of compatibility metrics.}
% Following HM~\citep{lyu2025test} and DAA~\citep{liudaa}, we use TA and CA as the primary metrics for real-time TTD evaluation, as they directly reflect online discovery quality and class consistency.
TA and CA directly reflect online discovery quality and class consistency as the primary metrics, while the four clustering metrics mainly serve compatibility with post-hoc NCD evaluation. The results in Table~\ref{tab:clumetric} demonstrate that our method also has better clustering performance on post evaluation.

\begin{figure}[t]
% \vspace{-10px}
    \centering    
    \includegraphics[width=1\linewidth]{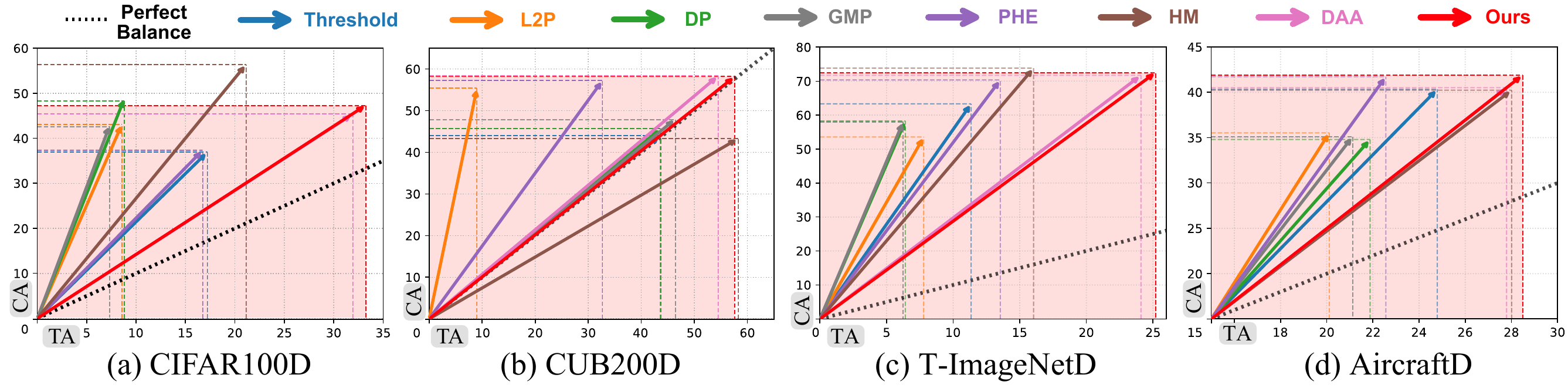}
    \vspace{-20px}
    \caption{
    Comparison on balance of Real-time TA and CA. Perfect Balance represents the most balanced TA and CA (TA = CA).
    }
    \label{fig:balance}
    \vspace{-10px}
\end{figure}

\noindent\textbf{Balance of TA \& CA}. 
TA and CA are two complementary indicators, and a high individual indicator does not necessarily indicate a good overall classification performance. Therefore, higher values of both metrics, along with greater balance between them, indicate better classification performance. We compared the balance in Fig.~\ref{fig:balance} and found that some methods have higher values but deviate far from the Perfect Balance line. And our method is more balanced and effective (larger red areas) on different datasets.

\begin{table}[t]
% \vspace{-10px}
\centering
\setlength{\tabcolsep}{4pt}
\caption{Comparisons of clustering metrics of post evaluation on four datasets.}
\vspace{-10px}
\resizebox{\linewidth}{!}{
\begin{tabular}{l|cccc|cccc|cccc|cccc}
\toprule
\multirow{2}{*}{Method} & \multicolumn{4}{c|}{\textbf{CIFAR100D}}                                & \multicolumn{4}{c|}{\textbf{CUB200D}}                                & \multicolumn{4}{c|}{\textbf{T-ImageNetD}}                                & \multicolumn{4}{c}{\textbf{AircraftD}}                                \\
                        & HCA & ARI & NMI & VM & HCA & ARI & NMI & VM & HCA & ARI & NMI & VM & HCA & ARI & NMI & VM \\
\midrule
HM~\citep{lyu2025test}      & 63.69 & 48.35 & 73.08 & 73.08 & 50.46 & 27.81 & 70.25 & 70.25 &  56.64 & 40.35 & 72.70 & 72.70  & 33.30 & 21.33 & 46.46 & 46.46 \\
DAA~\citep{liudaa}      & 65.62 & 53.96 & 75.20 & 75.21  & 58.19 & 32.43 & 75.29 & 75.29  & 58.91 & 42.50 & 73.53 & 73.53 & 32.08 & 21.20 & 46.44 & 46.44 \\
\rowcolor{orange!20} Ours      & 66.38 & 53.54 & 75.36 & 75.36 & 58.34 & 33.70 & 75.53 & 75.53 & 58.25 & 43.26 & 73.36 & 73.36 & 33.21 & 22.13 & 47.02 & 47.02 \\
\bottomrule
\end{tabular}
}
\label{tab:clumetric}
% \vspace{-5px}
\end{table}

%一排四个
\vspace{-5px}
\subsection{Analysis on Energy ring}
\vspace{-5px}
\noindent\textbf{Comparison on norm and confidence.} Fig.~\ref{fig:normmethod} shows that there is indeed a certain norm difference between the features of known and unknown classes obtained by the backbone network, as stated in~\citep{objectsphere, lyu2025test}. The average norm of unknown classes is slightly lower than that of known classes. Our method trains simulated classes as unknown classes and significantly reduce the norm of the simulation classes, which successfully simulate unknown classes to a certain extent, thus reducing the norm of the unknown classes. And our strategy makes their average confidence differences larger, which means that unknown classes are easier to be discovered.

% 以上这段分析中，要说明白我们ERPC策略通过设置模拟类别，起到了xx作用，最终扩大了已知类与未知类之间的差异。

%好的

%放到附录里

% \vspace{-10px}
% \begin{figure}[htbp]
%     \centering    
%     \includegraphics[width=\linewidth]{figs/normvs.pdf}
%     \vspace{-20px}
%     \caption{
%     Comparison on Norm and Confidence.
%     }
%     \label{fig:normvs}
%     \vspace{-15px}
% \end{figure}

% \begin{wraptable}[8]{r}{0.585\textwidth}
% \vspace{-35px}
% \centering
% % \small
% \setlength{\tabcolsep}{3pt}
% \caption{Different energy weight during testing.} 
% \vspace{-2px}
% \label{tab:norm weight}
% \resizebox{1\linewidth}{!}{
% \begin{tabular}{c|ccc|cccc}
%     \toprule
%     Energy weight & \multicolumn{3}{c|}{\textbf{Real-time Eval}} & \multicolumn{4}{c}{\textbf{Post Eval}}       \\ 
%     \cmidrule{2-8} 
%      $\alpha$  & TA            & CA    & T+C     & TA       & CA   & T+C  & KF    \\
%     \midrule
%         0 (No norm)             & 31.81 & 44.59 & 76.40 & 35.25 & 36.99 & 72.24 & 4.30 \\
%      0.01             &  32.56 & 45.58 & 78.14 & 36.42 & 37.43 & 73.85 & 4.07 \\
%      0.05             & 33.52 & 47.66 & 81.18 & 37.74 & 38.92 & 76.66 & 3.65  \\
%      0.10             &  29.25 & 48.23 & 77.48 & 34.92 & 40.95 & 75.87 & 3.83 \\
%      0.50             &  25.70 & 49.48 & 75.18 & 27.25 & 36.51 & 63.76 & 8.49  \\
%     (Norm-based)          & 12.87 & 27.16 & 40.06 & 20.92 & 24.79 & 45.71 & 12.90 \\
%     \bottomrule
% \end{tabular}
% }
% \vspace{-10pt}
% \end{wraptable}

\begin{table}[h]
\vspace{-10px}
	\centering
	\begin{minipage}{0.535\textwidth}
		\centering
		\small
		\makeatletter\def\@captype{table}\makeatother
		\setlength{\tabcolsep}{3pt}
\caption{Different energy weight during testing.} 
% \vspace{-10px}
\label{tab:norm weight}
\resizebox{1\linewidth}{!}{
\begin{tabular}{c|ccc|cccc}
    \toprule
    Energy weight & \multicolumn{3}{c|}{\textbf{Real-time Eval}} & \multicolumn{4}{c}{\textbf{Post Eval}}       \\ 
    \cmidrule{2-8} 
     $\alpha$  & TA            & CA    & T+C     & TA       & CA   & T+C  & KF    \\
    \midrule
        0 (No norm)             & 31.81 & 44.59 & 76.40 & 35.25 & 36.99 & 72.24 & 4.30 \\
     0.01             &  32.56 & 45.58 & 78.14 & 36.42 & 37.43 & 73.85 & 4.07 \\
     0.05             & 33.52 & 47.66 & 81.18 & 37.74 & 38.92 & 76.66 & 3.65  \\
     0.10             &  29.25 & 48.23 & 77.48 & 34.92 & 40.95 & 75.87 & 3.83 \\
     0.50             &  25.70 & 49.48 & 75.18 & 27.25 & 36.51 & 63.76 & 8.49  \\
    (Norm-based)          & 12.87 & 27.16 & 40.03 & 20.92 & 24.79 & 45.71 & 12.90 \\
    \bottomrule
\end{tabular}
}
	\end{minipage}
    \hfill
	\begin{minipage}{0.445\textwidth}
		\centering
		\small
		\makeatletter\def\@captype{table}\makeatother
        \setlength{\tabcolsep}{3pt}
    \centering
    \caption{Ablation study. T: Test-time Training, W: Energy-weighted prediction, P: Perturbation rectification, T+C: TA+CA.}
\vspace{-11px}
\label{tab:ablation}
\resizebox{1\linewidth}{!}{
\begin{tabular}{ccc|ccc|cccc}
    \toprule
    \multirow{2}{*}{\;T\;}  & \multirow{2}{*}{\;W\;}  & \multirow{2}{*}{\;P\;} & \multicolumn{3}{c|}{\textbf{Real-time Eval}}                                                                         & \multicolumn{4}{c}{\textbf{Post Eval}}                                                                                                           \\ \cmidrule{4-10} 
        &       &    &  TA            & CA    & T+C        & TA      & CA    & T+C   & KF                        \\
    \midrule
                                             
                        &                &     & 30.81 & 42.13 & 72.94 & 34.91 & 38.53 & 73.44 & 2.27   \\
 $\checkmark$ &       &   & 24.68 & 53.74 & 78.42 & 24.89 & 42.68 & 67.57 & 6.84  \\
&  $\checkmark$ &   & 29.61 & 45.54 & 75.15 & 32.44 & 36.45 & 68.89 & 3.05   \\

    $\checkmark$       & $\checkmark$ &  & 31.60 & 44.99 & 76.59 & 34.52 & 36.52 & 71.04 & 4.22  \\
    $\checkmark$       & $\checkmark$ & $\checkmark$ & 33.52 & 47.66 & 81.18 & 37.74 & 38.92 & 76.66 & 3.65  \\
    \bottomrule
\end{tabular}
}
    \label{tab:theta}
	\end{minipage}
% \vspace{-10px}
\end{table}

\noindent\textbf{Energy weight.}
We exploit the spatial variation in feature responses to re-weight categorical similarity, using a norm-based modulation that remains deliberately light~(Sec.\ref{subsec:joint}). The feature-to-category similarities already lie in [0, 1] and the overlap zone clusters around 0.7. Consequently, as Table~\ref{tab:norm weight} shows, only a small modulation factor is required, once the exponent becomes large, the relative reliabilities are overturned and the norm effectively seizes control of the decision; in the limiting case of a purely norm-governed weight, the accuracy collapses.

\noindent\textbf{Ablation study}.
Table~\ref{tab:ablation} summarizes the ablation results. Test-time Training (T) alone lifts CA from 42.13 to 53.74, but results in larger KF and lowers post-TA, revealing instability. Energy-weighted prediction (W) without T already outperforms the fixed model, pushing real-time TA+CA from 72.94 to 75.15, confirming the effectiveness of energy-weight assistance. Injecting perturbation consistency rectification (P) into the full pipeline helps rectify unstable samples, attaining the best overall trade-off (TA+CA 81.18 real-time and 76.66 post). These results demonstrate that W and P can effectively enable positive test-time update without catastrophic forgetting.

\begin{figure}[t]
% \vspace{-10px}
	\centering
	\begin{minipage}{0.49\textwidth}
		\centering
		\includegraphics[width=1\linewidth]{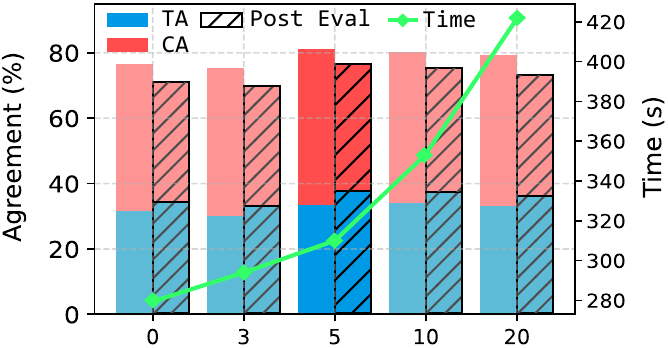}
		\vspace{-19px}
        \makeatletter\def\@captype{figure}\makeatother
        
		\caption{Different perturbations $k$ during testing.} 
		\label{fig:disturb}
	\end{minipage}
    \hfill
	\begin{minipage}{0.485\textwidth}
		\centering
		\includegraphics[width=1\linewidth]{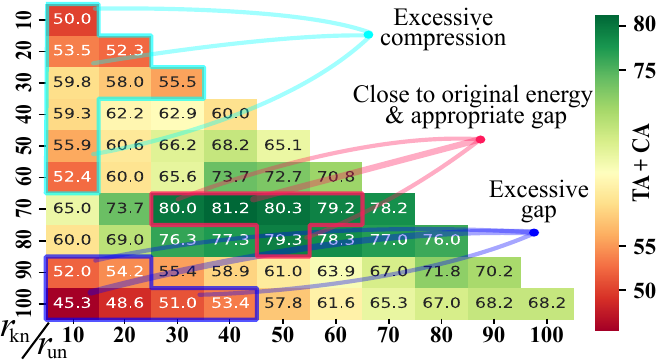}
		\vspace{-21px}
        \makeatletter\def\@captype{figure}\makeatother
        \caption{Different $r_\text{kn}$ \& $r_\text{un}$ during Warm-Up.}
        \label{fig:rkru}
	\end{minipage}
% \vspace{-21px}
\end{figure}

\noindent\textbf{Different Perturbation}.
We mainly enhance the stability among unknown classes through perturbations, and during testing, we make corrections based on the results of perturbations. In Fig.~\ref{fig:disturb}, insufficient perturbations makes the judgment of stability reckless and coincidental changes or no changes can be considered unstable or very stable, so the effect cannot be effectively improved, while excessive perturbations does not significantly improve the effect and introduces greater time cost. 

\noindent\textbf{Energy rings during warm-up.} We treat the energy ring $r_\text{kn}$ and $r_\text{un}$ as a coupled hyper-parameter pair that determines the spatial energy layout of feature. Fig.~\ref{fig:rkru} shows  when $r_\text{un}$ is too small the inner ring is pushed close to the origin with a large gap between $r_\text{kn}$, making classification difficult and yielding unstable clusters. The experiment of Fig~\ref{fig:normmethod} finds that the norm of backbone features for known classes centers around 70. Therefore, we fix the known class feature norm at 70 and handle unknown classes for better results.  

\begin{figure*}[h]
% \vspace{-10px}
    \centering    
    \includegraphics[width=0.95\linewidth]{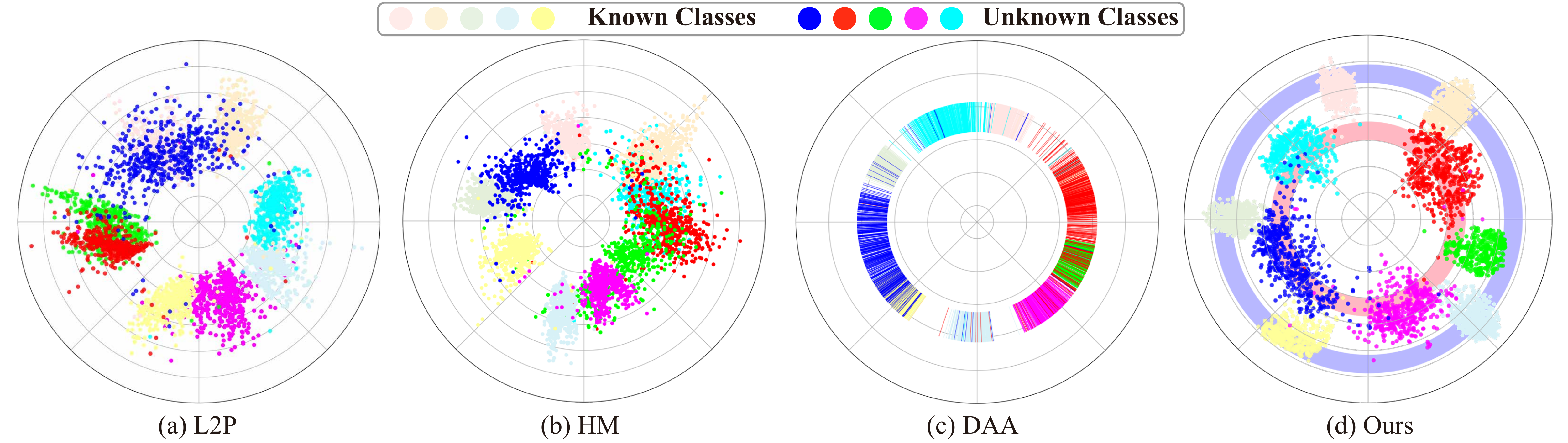}
    % \vspace{-10px}
    \caption{
    T-SNE visualization in angular space on CIFAR100D with 5 known and 5 unknown classes.
    }
    \label{fig:tsne}
    % \vspace{-18px}
\end{figure*}

\noindent\textbf{Post visualization using T-SNE}.
In Fig.~\ref{fig:tsne}, we employ t-SNE~\citep{van2008visualizing} to visualize the distribution of test samples. Methods that directly update the base model parameters often risk disrupting the model's existing structure, leading to destructive confusion between classes. In contrast, HM methods cannot update the model, leaving the feature distribution static. Meanwhile, DAA dismisses the spatial difference of feature, limiting its angular amplification effect. In contrast, our method strikes a balance by not only distinguishing features based on angles but also accounting for the energy intensity between the inner and outer energy rings, and by learning updates that separate features across different classes and maintain clearer spatial boundaries than other methods.

\begin{figure*}[h]
% \vspace{-15px}
    \centering    
    \includegraphics[width=0.95\linewidth]{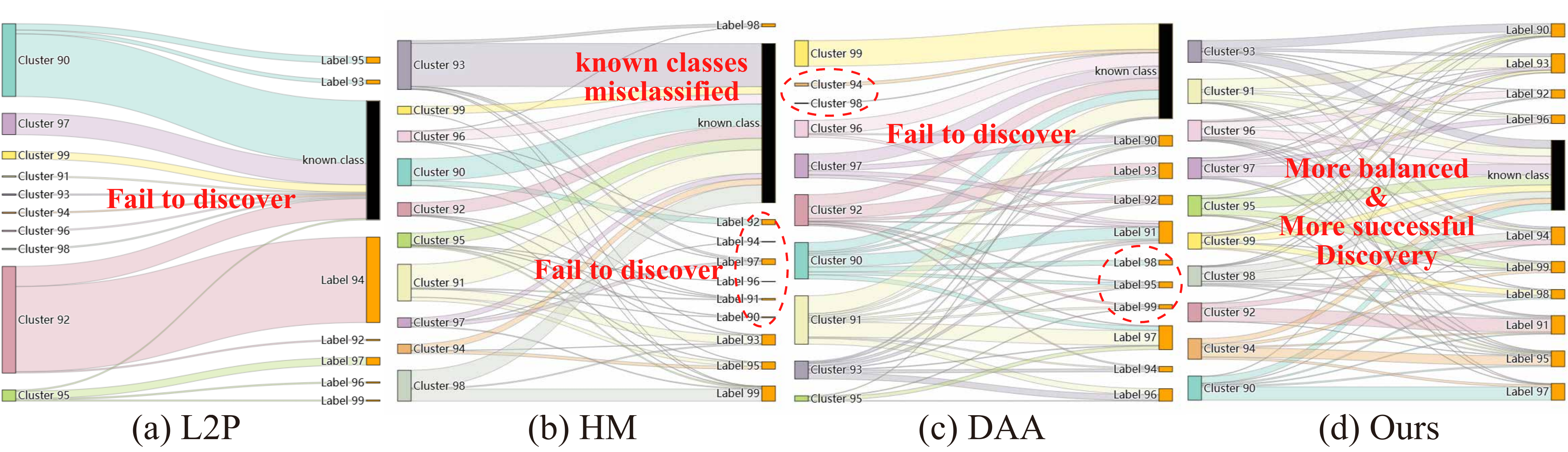}
    \vspace{-5px}
    \caption{
    Matching comparison between unknown class predictions and ground truth across methods on CIFAR-100D (90+10). Black rectangles indicate old class samples misclassified as unknowns.
    }
    \label{fig:sankey}
    % \vspace{-10px}
\end{figure*}

\noindent\textbf{Comparison of unknown class label matching.}
Fig.~\ref{fig:sankey} shows the matching between predictions and ground truth, where the base-model updating method easily leads to chaotic predictions and misclassification. In the fixed-energy method like HM, lots of old samples are classified into unknown clusters. DAA has improved its ability to discover new classes, but there are still some categories that fail to be discovered. Our method can discover a larger number of new classes and make the matching more balanced while classifying fewer known classes as unknown.

\section{Conclusion}
In this paper, we introduced the Energy Ring framework to address the challenging Test-Time Discovery task. Our method enables instant recognition of unknown classes while safeguarding the accuracy of previously learned ones. The lightweight energy ring adapter compresses feature onto concentric energy rings, preserves known-class directions, enlarges the energy gap and produces compact, stable clusters for unknown classes. Comprehensive experiments on four datasets demonstrate that Energy Ring outperforms in both adaptability and stability. 
Although energy ring adds marginal computation during warm-up and test-time training, the overhead is negligible compared with the gains in discovery performance. However, the current energy cycle still needs to rely on pre-set energy tracks. Future work will explore more efficient adapter designs, noise-robust update rules, and extensions to multi-modal and continual learning scenarios. And we will explore how to dynamically set energy tracks to achieve wider applicability in the open world.

% \newpage

% \subsubsection*{Author Contributions}
% If you'd like to, you may include  a section for author contributions as is done
% in many journals. This is optional and at the discretion of the authors.

% \subsubsection*{Acknowledgments}
% Use unnumbered third level headings for the acknowledgments. All
% acknowledgments, including those to funding agencies, go at the end of the paper.

\bibliography{main}
\bibliographystyle{bibstyle}

% \newpage
% \appendix
% \section*{Appendix}
% \input{sec/X_suppl}

\end{document}

%% file: math_commands.tex
\usepackage{amsmath,amsfonts,bm}

\def\eqref#1{equation~\ref{#1}}
\def\1{\bm{1}}

\DeclareMathAlphabet{\mathsfit}{\encodingdefault}{\sfdefault}{m}{sl}
\SetMathAlphabet{\mathsfit}{bold}{\encodingdefault}{\sfdefault}{bx}{n}